\documentclass[review]{elsarticle}
\usepackage{hyperref}
\usepackage{hyperref}
\usepackage{amsmath}
\usepackage{colortbl}
\usepackage{algorithmic}
\usepackage{algorithm}
\usepackage{listings}
\usepackage{multirow}
\usepackage{makecell}
\journal{Journal of \LaTeX\ Templates}

\newlength\savewidth\newcommand\shline{\noalign{\global\savewidth\arrayrulewidth
  \global\arrayrulewidth 1pt}\hline\noalign{\global\arrayrulewidth\savewidth}}
  
\def\eg{\emph{e.g.}} 
\def\ie{\emph{i.e.}} 
\def\etal{\emph{et~al.}} 
\begin{document}

\begin{frontmatter}

\title{Multiple-environment Self-adaptive Network \\ for Aerial-view Geo-localization}

\author[mymainaddress]{Tingyu Wang}
\ead{wongtyu@hdu.edu.cn}

\author[thirdaddress]{Zhedong Zheng}
\ead{zhedongzheng@um.edu.mo}

\author[mymainaddress,mysecondaryaddress]{Yaoqi Sun}
\ead{syq@hdu.edu.cn}

\author[mymainaddress]{Chenggang Yan\corref{mycorrespondingauthor}}
\cortext[mycorrespondingauthor]{Corresponding author}
\ead{cgyan@hdu.edu.cn}

\author[fourthaddress]{Yi Yang}
\ead{yangyics@zju.edu.cn}

\author[thirdaddress]{Tat-Seng Chua}
\ead{chuats@comp.nus.edu.sg}

\address[mymainaddress]{Intelligent Information Processing Lab, Hangzhou Dianzi University, China}
\address[mysecondaryaddress]{Lishui Institute of Hangzhou Dianzi University, China}
\address[thirdaddress]{Faculty of Science and Technology, and Institute of Collaborative Innovation, University of Macau, China}
\address[fourthaddress]{College of Computer Science and Technology, Zhejiang University, China}
\begin{abstract}
Aerial-view geo-localization tends to determine an unknown position through matching the drone-view image with the geo-tagged satellite-view image. This task is mostly regarded as an image retrieval problem. The key underpinning this task is to design a series of deep neural networks to learn discriminative image descriptors. However, existing methods meet large performance drops under realistic weather, such as rain and fog, since they do not take the domain shift between the training data and multiple test environments into consideration. 
To minor this domain gap, we propose a Multiple-environment Self-adaptive Network (MuSe-Net) to dynamically adjust the domain shift caused by environmental changing. 
In particular, MuSe-Net employs a two-branch neural network containing one multiple-environment style extraction network and one self-adaptive feature extraction network. As the name implies, the multiple-environment style extraction network is to extract the environment-related style information, while the self-adaptive feature extraction network utilizes an adaptive modulation module to dynamically minimize the environment-related style gap. Extensive experiments on three widely-used benchmarks, \ie, University-1652, SUES-200, and CVUSA, demonstrate that the proposed MuSe-Net achieves a competitive result for geo-localization in multiple environments. Furthermore, we observe that the proposed method also shows great potential to the unseen extreme weather, such as mixing the fog, rain and snow.
\end{abstract}


\end{frontmatter}

% \linenumbers
\begin{figure}[htbp]
  \centering
  \includegraphics[width=0.9\linewidth]{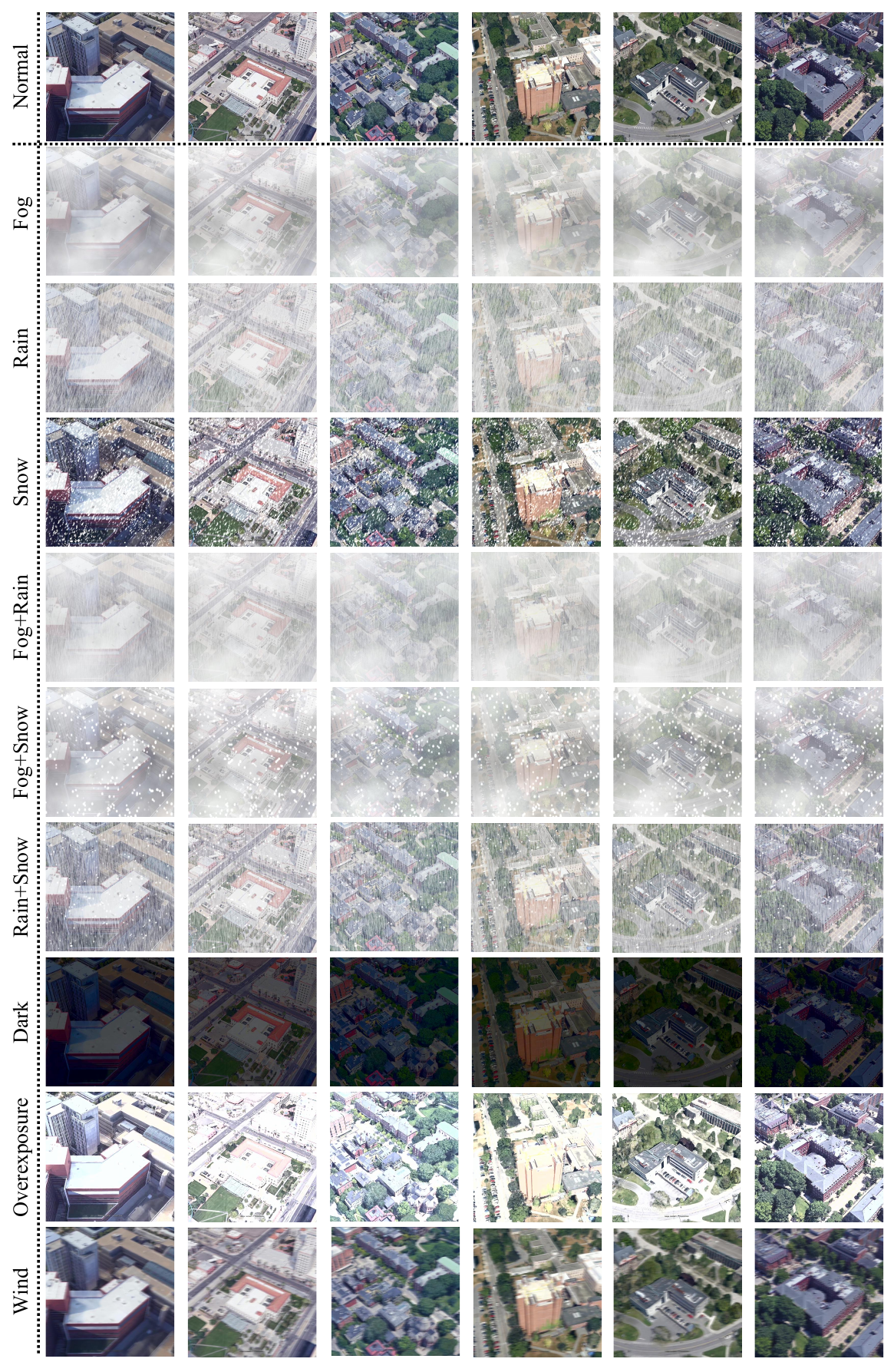}
%   \vspace{-.1in}
  \caption{Examples of synthesized environments on University-1652~\cite{zheng_university-1652_nodate}, which raise challenges on the robustness of the drone vision. Specifically, we generate these images by adding different environmental styles into the normal drone-view images. 
  Each column and row corresponds to different geographical locations and environments.}
  \label{fig:1}
\end{figure}
\section{Introduction}
Aerial-view geo-localization generally refers to retrieving the corresponding images between drone and satellite platforms. It recently could be deployed to many fields, such as drone navigation, event detection, aerial photography, and so on~\cite{zheng_university-1652_nodate,wang2021each,lin2022joint}. In application, given a drone-view image as the query, the retrieval system intends to search the most relevant satellite image from the candidate gallery. The satellite-view images possess geo-tag information, such as GPS. Thus the drone can naturally determine its geographic location. Compared with ground-view images with more occlusion, \eg{}, tree, drone-view images support excellent visibility. However, aerial-view geo-localization remains challenging due to the cross-view domain shift caused by the viewpoint and the environment.

Convolutional neural networks (CNNs) have recently received primary support in aerial-view geo-localization due to the strong potential to learn invariant image representations. Relying on CNNs, a two-branch network~\cite{liu_lending_2019, shi_optimal_nodate} as the prototype has been widely employed in related studies. The metric learning~\cite{Shi_2020_CVPR, hu_cvm-net_2018} and the classification loss~\cite{zheng_university-1652_nodate, wang2021each} are two predominant choices to optimize this prototype. A number of extension works adjust the spatial layout of image 
semantics~\cite{shi_spatial-aware_nodate}
% semantics~\cite{shi_spatial-aware_nodate, fu2019sta} 
to extract view-invariant features or deploy part-based matching~\cite{wang2021each} to roughly align local information. All these existing methods concentrate on mitigating the cross-view domain gap introduced by viewpoint change. One scientific problem raises: how can the model cope with the environmental domain shift? Recently, some researchers~\cite{wu2019ace} 
% researchers~\cite{hendrycks2018benchmarking, wu2019ace} 
point out that a trained network is easy to collapse for unfamiliar input distributions. As a result, existing networks are likely to fail in multi-environment inference of geo-localization. Our research does not focus on improving the general matching accuracy. In contrast, we mainly study the robustness of current methods on cross-platform media against different environments. Specifically, we aim to relieve the negative impact of geo-localization in noisy environments, such as rain and fog. This robustness task is also non-trivial since the drone can easily encounter environmental changes during flight. It is well known that bad weather can lead to invisibility of geographic targets and even cause serious flight accidents~\cite{investigation}. Therefore, employing the aerial-view geo-localization system to retrieve in multiple environments is a meaningful and practical topic. This topic touches on domain generalization (DG)~\cite{li2017deeper}, and a direct effort is to let the geo-localization model 'remember' the distribution of a location in different environments during training. However, many studies~\cite{li2017deeper,chattopadhyay2020learning} in domain generalization have demonstrated that forcing the entire model or features to be domain invariant is challenging. We know that humans recognize a previously seen location again by eliminating the interference caused by environmental changes rather than remembering how the location looked under different environments. This human cognitive mechanism inspires our research. Specifically, when testing, we hope that the trained drone-to-satellite geo-localization system can adaptively filter out the domain shift caused by environmental changes. While achieving this goal is non-trivial, two issues need to be discussed.
One is the reproduction of the environmental style information. We consider a reasonable assumption that the seen environments can replay in a new scenario. Therefore, we need the system to be able to independently reproduce different environmental style information from the inputs during testing. Then the style information is utilized to balance the environmental domain shift. To do so, we expand the training data for environmental diversity first. The data collection is expensive and difficult since one fixed position calls for images shot in different weather. Another reason is that unlike the acquisition of ground-level images, the collection of drone-view images needs to be operated by professionals. Generative adversarial networks (GANs)~\cite{2014generative} and data augmentation are two appealing choices to replace manual dataset production. GANs recently can synthesize images that are not only of high quality but also possess sufficient diversity. However, the key of our research is not to generate stylized images with high quality. Considering the speed and flexibility, we abandon GANs and choose an off-the-shelf image-based style transformation library~\cite{imgaug} to pre-process images. After processing images, we obtain nine synthetic environmental images for one geographical location, \ie{}, fog, rain, snow, fog add rain, fog add snow, rain add snow, dark, overexposure, and wind (see Figure \ref{fig:1}). We then require the system to be able to extract different environment knowledge to achieve reproduction. In vanilla CNNs, the image information contained in extracted features is tangled. Some methods~\cite{ilse2020diva,chattopadhyay2020learning} suggest that representations can be disentangled into different parts using specific domain labels or domain-related prior knowledge. We follow this idea and encode the environmental information from input images by supervised learning, where the environmental labels are automatically annotated during the image transformation. 
Another issue is how to utilize the style information to minimize the environmental style gap. We suppose that the identity information is domain-agnostic and the environmental style is domain-specific for one location. Some reference methods retain only the domain-agnostic information for downstream visual tasks~\cite{khosla2012undoing,li2017deeper} or employ instance normalization~\cite{ulyanov2017improved} to resist the effect of image style changes~\cite{pan2018two}. Unlike these methods, our system aims to align the domain-specific distribution on the fly. To achieve this goal, we first borrow experience from IBN-Net~\cite{pan2018two} to process images and obtain visual features. IBN-Net integrates batch normalization (BN)~\cite{ioffe2015batch} and instance normalization (IN) into residual blocks of shallow layers. BN is used to retain the discrimination of features~\cite{huang2017densely,he2016deep}, and IN is employed to filter out domain-specific style information from the content~\cite{pan2018two}. However, IN applies the same treatment to multiple styles of content. To satisfy the dynamic adaptability, we further introduce spatially-adaptive denormalization (SPADE)~\cite{park2019semantic} and integrate SPADE into a residual structure, called Residual SPADE, in our system. Same as SPADE, Residual SPADE is a conditional normalization module that allows flexible modulation of image styles by scales and biases learned from the external data.
In light of the above analysis, we propose a two-branch learning framework called Multiple-environment Self-adaptive Network (MuSe-Net). The design of MuSe-Net is based on IBN-Net yet with more scalability. In particular, we insert Residual SPADE after the instance normalization layer of IBN-Net as the self-adaptive feature extraction network to construct one branch of MuSe-Net. The other branch of MuSe-Net is a multiple-environment style extraction network, which parameterizes the environment information as inputs of Residual SPADE. Two branches have the same inputs, ensuring that the self-adaptive feature extraction network can utilize the corresponding environmental information from the style extraction network to dynamically close the environmental style gap. 
\par 
The main contributions of this work are summarized as follows.
\begin{itemize}
\item \textbf{We identify one key challenge},~\ie, the weather and illumination changes, when applying the visual geo-localization system to the real-world scenario. The large visual changes usually compromise the reliability of the existing methods. To address this limitation, based on simulated multi-weather data, \textbf{we adopt an adaptive adjustment strategy of style information} an present an end-to-end learning framework called Multiple-environment Self-adaptive Network (MuSe-Net). MuSe-Net applies a dual-path CNN model to extract the environment-related style information and dynamically minimize the environment-related style gap, such as weather and light changes. To motivate the model to learn discriminative features, \textbf{we further introduce a module called Residual SPADE}, which utilizes the residual structure to optimize the training of MuSe-Net.

\item Extensive experiments on three prevailing multi-platform geo-localization benchmark, \ie, University-1652~\cite{zheng_university-1652_nodate}, SUES-200~\cite{zhu2023sues}, and CVUSA~\cite{zhai_predicting_2017}, show that our method achieves superior results for geo-localization in multiple environments. Meanwhile, for an unseen extreme weather, \ie, mixing the fog, rain and snow, MuSe-Net still arrives at competitive results.
\end{itemize} 

\section{Related Work} 
In this section, we briefly discuss the relevant works in two aspects, including CNN-based cross-view geo-localization and domain generalization.
\subsection{CNN-based Cross-view Geo-localization}
Existing cross-view geo-location methods mostly focus on solving the visual gap caused by the changing appearance in different viewpoints. In order to gain the discriminative image representation, some pioneering works~\cite{lin2013cross}
% some pioneering works \cite{SemanticCM, lin2013cross} 
make many efforts on hand-crafted feature matching. Due to the powerful capability on image representation~\cite{russakovsky2015imagenet}, deep convolutional neural networks (CNNs) become the prevalent choice for feature extraction. Follow this line, Workman~\etal{}~\cite{workman_location_2015} first attempt to employ an AlexNet~\cite{krizhevsky2012imagenet} pre-trained on Imagenet~\cite{russakovsky2015imagenet} and Places~\cite{zhou2014learning} to extract deep features for cross-view geo-localization. They prove that the top layers of CNN include rich information of geographic location. Further, Workman~\etal{}~\cite{workman_wide-area_2015} extend their work by minimizing the distance of cross-view features and gaining improved performance. Lin~\etal{}~\cite{lin_learning_2015} define the matching task as similar to the face verification, and deploy the contrastive loss~\cite{hadsell2006dimensionality} to optimize a modified Siamese Network~\cite{chopra2005learning}. Later on, Hu~\etal{}~\cite{hu_cvm-net_2018} insert a NetVLAD into the high-level layer of a Siamese-like architecture to aggregate the local feature, which facilitates image descriptors against the viewpoint changes. 
Considering the importance of orientation, Liu~\etal{}~\cite{liu_lending_2019} encode the corresponding coordinate information into the network for the discriminative feature. Tian~\etal~\cite{tian2020cross} propose an orientation normalization network to alleviate the effect of the variation of orientation. Wang~\etal~\cite{wang2021each} design a square-ring partition strategy to cope with the image rotation. Discussing on a limited Field of View (FoV) setting, DSM~\cite{Shi_2020_CVPR} provides a dynamic similarity matching module to align the orientation of cross-view images.
In order to align the spatial layout information,  Zhai~\etal{}~\cite{zhai_predicting_2017} encode and transfer the semantic information of ground images to aerial images. Regmi and Shah~\cite{Regmi_2019_ICCV} apply a generative model to synthesize an aerial image from a panoramic image of the ground. Shi~\etal{}~\cite{shi_optimal_nodate} resort to the optimal transport theory to compare and adjust pairwise image distribution in the feature level. Another work of Shi~\etal{}~\cite{shi_spatial-aware_nodate} directly utilize the polar transform to accomplish the pixel-level alignment of semantic information between ground images and satellite images. Lin~\etal{}~\cite{lin2022joint} exploit keypoints to enrich the model capability of learning robust features against viewpoints. Dai~\etal{}~\cite{dai2022fsra} employ the heat distribution of the feature map to find and align critical regions in images from different viewpoints.
In the aspect of optimizing different training objectives, Vo and Hays~\cite{vo_localizing_2017} investigate a variety of CNN architectures for cross-view matching and gain the best performance through employing a soft margin triplet loss to optimize a triplet CNN. Hu~\etal{}~\cite{hu_cvm-net_2018} design a weighted soft-margin ranking loss that speeds up the convergence rate. Cai~\etal{}~\cite{Siam-FCANet} introduce a hard exemplar reweighting triplet loss to improve the retrieval. Zheng~\etal{}~\cite{zheng_university-1652_nodate} imitate the classification tasks and use instance loss~\cite{lin2019improving,zheng2020dual} 
as the proxy targets to solve the cross-view image retrieval. Sun~\etal~\cite{sun2023f3} treat the multiview features as probability distributions and eliminate the feature differences by constraining the transmission loss. 
There are also some works~\cite{Siam-FCANet,shi_spatial-aware_nodate} that employ the attention mechanism to locate the interesting areas, which effectively promotes the discrimination of features. 

\textbf{In contrast to existing works, we study a new real-world scenario where the drone may encounter different weather and illumination, which leads to the same location with multiple domain distributions.} For this multiple domain problem, we explicitly parameterize environmental information to align the domain distribution. Therefore, the drone is able to localize unseen positions in previously seen environments.

\subsection{Domain Generalization}
Generally, domain generalization (DG) is mentioned in multi-source domain problem. Similar to the topic of domain adaptation (DA), domain generalization tends to address the domain shift introduced by the statistical differences of multiple domains. However, compared with domain adaptation that employs the labeled or unlabeled data of target domain during training, domain generalization concentrates on only leveraging the multi-source data to learn robust data representations which could be potentially useful in different marginal distributions,~\eg{}, unseen target domains. 

There are many strategies that attempt to achieve domain generalization. Some researchers argue that minimizing various well-designed distribution metrics~\cite{chen2023domain}
% metrics~\cite{li2018domain,li2020domain,chen2023domain}
can realize multi-domain alignment. The statistical learning theory~\cite{vapnik1999nature} suggests that the diversity of training samples can boost the generalization of learning models. Therefore, other researchers propose advanced augmentation algorithms~\cite{zhang2018mixup}
% algorithms~\cite{zhou2021domain,zhang2018mixup}
to lessen the domain gap. Adversarial approaches~\cite{rahman2020correlation} rely on confusing a domain discriminator to learn domain-invariant features, which can also alleviate the domain shift. Meta-learning~\cite{zhang2021curriculum} enables the model to learn new concepts and skills fast with a few training examples. Based on meta-learning, MAML~\cite{finn2017model} gains great success in domain generalization, and some improved MAML frameworks are also subsequently presented in~\cite{zhang2022lsrml}. 
% in~\cite{zhang2022generalizable,zhang2022lsrml}. 
Disentangled representation learning intends to learn the 
common and exclusive information from multiple domains and then processes these information separately to obtain robust features for future predictions. According to that, one group of approaches considers the disentanglement at the feature level. Khosla~\etal{}~\cite{khosla2012undoing} decompose a classifier based on the SVM into common vectors and bias vectors, and abandon the bias part when applying in unseen domains. Later, Li~\etal{}~\cite{li2017deeper} employ the neural network to re-implement this concept. DMG~\cite{chattopadhyay2020learning} adopts a learnable mask to select and balance the domain-invariant and domain-specific features and demonstrates that analysing domain-specific components can assist to the prediction at test-time. Another solution supports encoding the multiple knowledge into different latent spaces.
Ilse~\etal{}~\cite{ilse2020diva} provide a VAE~\cite{kingma2013auto}-type framework to learn three complementary sub-spaces for domain-invariant classification. There is also literature documenting the use of GANs~\cite{zheng2019joint} to construct two latent spaces: one for identity confirmation and the other serving the domain-related information.

\section{Proposed Method}
\begin{figure*}[htbp]
  \centering
  \includegraphics[width=1\linewidth]{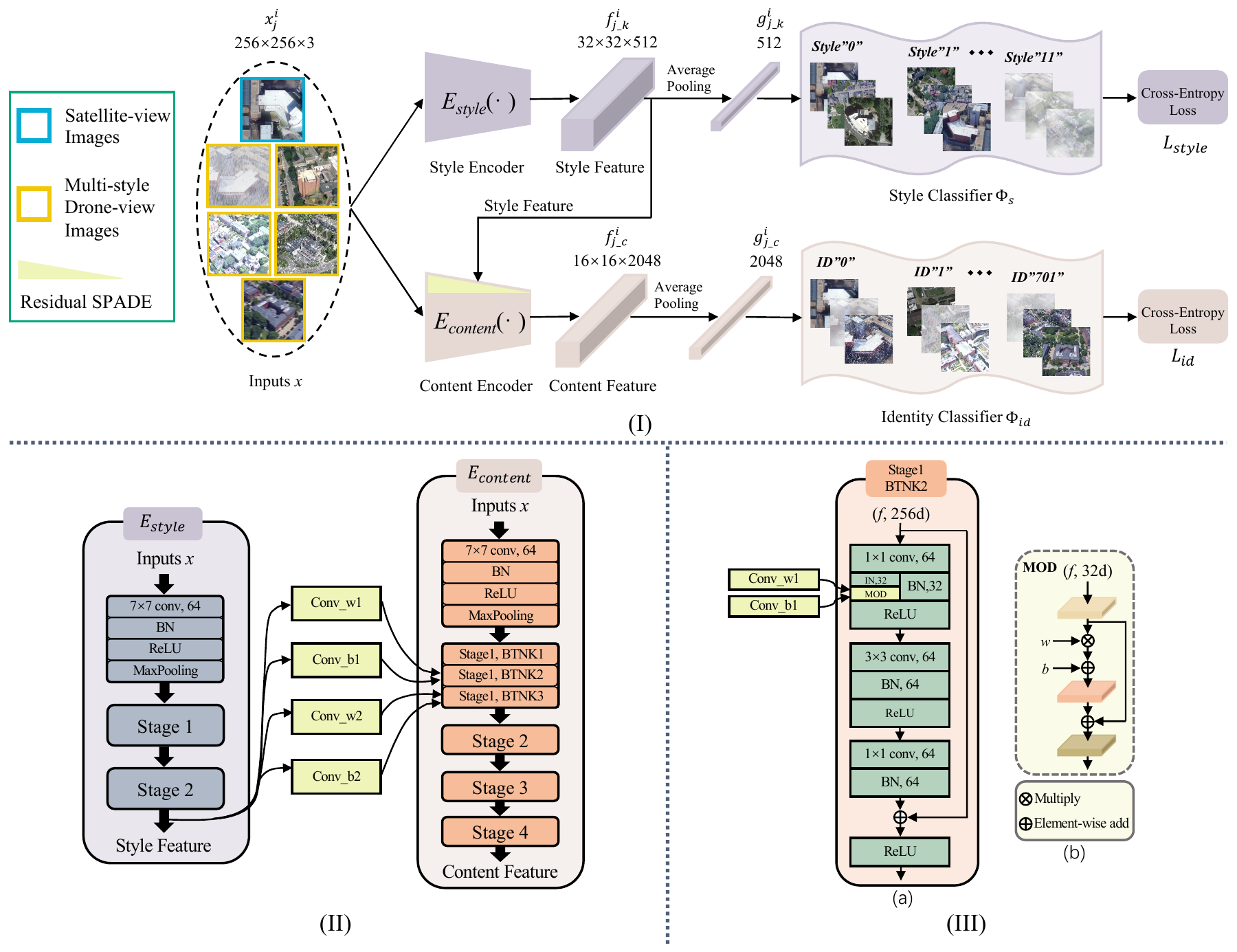}
%   \vspace{-.1in}
  \caption{(I) A schematic overview of MuSe-Net. One batch of inputs contains the same number of satellite and drone images, and the style of satellite images is invariable. MuSe-Net consists of two branches. Purple blocks indicate the multiple-environment style extraction branch (\textcolor[RGB]{172,165,182}{purple branch}), and pink blocks denote the self-adaptive feature extraction branch (\textcolor[RGB]{190,176,171}{pink branch}). The purple branch is employed to group the same style features. Then style features extracted from the style encoder are fed into Residual SPADE. Residual SPADE is embedded into the content encoder, which belongs to the pink branch. The pink branch is applied to narrow down the distance of inputs with the same geo-tag. (II) Detailed demonstration of information interaction between the style encoder and the content encoder. The extracted style information is first convolved to produce the modulation parameters in Residual SPADE. After that, we utilize the learned modulation parameters to modulate middle features of inputs in the content encoder. (III) Illustration of the location of Residual SPADE in one bottleneck of the content encoder (a) and the calculation flow of the modulation operation (b). In Residual SPADE, a group of modulation parameters $w$ and $b$ comes from two convolutional layers, \ie, Conv\_w1 and Conv\_b1. Afterwards the learned $w$ and $b$ are applied to modulate the activation of instance normalization (IN). 
  }
  \label{fig:2}
\end{figure*}

In this paper, we consider a more practical problem where the drone could encounter multiple environments, which cause the domain shift and drop the performance of aerial-view geo-localization. We challenge the problem by proposing a Multiple-environment Self-adaptive Network (MuSe-Net) (see Figure~\ref{fig:2}). In Section~\ref{section:1}, we first provide the problem definition and notations. We then revisit the relevant technologies of our method in Section~\ref{section:2}. Finally, we detail MuSe-Net in Section~\ref{section:3}.
\subsection{Problem Definition and Notations}\label{section:1}
The multiple environment aerial-view geo-localization task assumes the scenario that satellite-view images are constant while the style of drone-view images is variable with environmental changes. The different environmental information induces the existing methods which are difficult to search geographic target images with the same identities between different viewpoints. Our research focuses on reducing the interference of environmental styles when retrieving between two aerial viewpoints. Let $\mathcal{X} = \left\{x^i\right\}_{i=1}^N$,
% $\mathcal{Y}_{id} = \left\{y_{id}^i\right\}_{i=1}^N$ 
$\mathcal{Y}_{ID} = \left\{y_{ID}^i\right\}_{i=1}^N$ be the inputs and identity labels, respectively, for one multiple environmental aerial-view geo-localization dataset. $N$ is the number of images, and $y_{ID} \in [1, C]$, and $C$ indicates the number of identities. Conventional, the inputs $\mathcal{X}$ consist of $j$ domains, and $j \in \{1,2\}$. $\mathcal{X}_{1}$ denotes the satellite-view domain and $\mathcal{X}_{2}$ denotes the drone-view domain. These two domains share the same identity labels $\mathcal{Y}_{ID}$. In our setting, we follow the previous definition and add one style space. In particular, holding $\mathcal{Y}_{ID}$ constant, we expand the original drone-view domain $\mathcal{X}_{2}$ to multi-environment drone-view domain $\mathcal{X}_{2\_k}$. The subscript $k \in [1, K]$, and $K$ indicates the number of environmental styles. Since the style of the satellite-view domain is constant, we denote this domain as $\mathcal{X}_{1\_k}$ and $k = 0$. Therefore, the added style label $\mathcal{Y}_{style} = \left\{y_{style}^i\right\}_{i=1}^N$ includes $K+1$ styles, \ie{}, $y_{style} \in [0, K]$.  

\subsection{Relevant Technologies Revisit}\label{section:2}
% \textbf{IBN-Net~\cite{pan2018two}.} Batch normalization (BN) has recently been a basic component of CNNs. BN maintains the discrimination of features by utilizing the global statistics (\ie{}, mean and variance) recorded in training to normalize the testing sample. However, global statistics contains domain-specific knowledge (\eg, style information) inevitably. In domain generalization, the domain-specific information induces the domain shift, which causes a trained model in one domain with poor performance in another domain. For instance, in scene parsing, due to the different image styles in two datasets, a CNN-based model trained on Cityscapes~\cite{cordts2016cityscapes} exhibits notable performance drops when testing on GTA5~\cite{richter2016playing}, even though two datasets include similar semantics. In contrast with BN, instance normalization (IN) discards the global statistics. With the learned affine parameters, IN intends to close the style gap between each testing and training sample. Therefore, IN resists the effect of the style discrepancy but damages the discrimination simultaneously. Considering the advantages of BN and IN, IBN-Net integrates IN and BN as building blocks to extract style-invariable features and achieves a competitive result in the cross-domain scene parsing task.
\textbf{IBN-Net~\cite{pan2018two}}. Batch normalization (BN) maintains the discrimination of features by utilizing the global statistics (\ie{}, mean and variance) recorded in training to normalize the testing sample. Instance normalization (IN) discards the global statistics. With the learned affine parameters, IN intends to close the style gap between each testing and training sample. Therefore, IN resists the effect of the style discrepancy but damages the discrimination simultaneously. Considering the advantages of BN and IN, IBN-Net integrates IN and BN as building blocks to extract style-invariable features and achieves a competitive result in the cross-domain scene parsing task.

\textbf{Spatially-adaptive denormalization (SPADE)~\cite{park2019semantic}.} SPADE is a conditional normalization module that first requires external data to generate the learned affine parameters. Then the normalized activations are modulated by the learned affine parameters. SPADE can be simplified formulated as: 
\begin{align}\label{spade}
    SPADE(u,v) &= \sigma(v) \cdot \frac{u-\mu(u)}{\sigma(u)} + \mu(v),
\end{align}
where $u$ is the input feature and $v$ is the corresponding style feature. $\mu(u)$ and $\sigma(u)$ compute the mean and variance of feature $u$, respectively. $\sigma(v)$ and $\mu(v)$ are the learned scale and bias to modulate the normalized feature $u$. 

\textbf{Discussion.}
Instance normalization (IN) plays a crucial role in improving the generalization capability of IBN-Net. However, as mentioned above, IN dilutes the content discrimination. Unlike IBN-Net trained on Cityscapes, our approach deploys multiple environmental style images to train MuSe-Net. When there are significant discrepancies between these styles, we suggest that IN could diminish more useful information in order to learn compromise parameters for style alignment of features. Therefore, we propose an \textbf{adaptive adjustment strategy of style information} which inserts the Spatially-adaptive denormalization (SPADE) after IN and employs the style information of images as the external condition to dynamically adjust the activation of IN. 
It is worth noting that we do not directly follow Equation~\ref{spade} to modulate the normalized activations. We reserve the affine parameters of IN and apply a residual structure to adjust the feature. We call the modified SPADE as \textbf{Residual SPADE}, which can be formulated as:
\begin{align}\label{spade_residual}
    R\_SPADE(u,v) &= \sigma(v) \cdot IN(u)  + \mu(v) + IN(u) \notag\\
                   &= IN(u) \cdot (1 + \sigma(v)) + \mu(v),
\end{align}
where $R\_SPADE$ denotes Residual SPADE. Characters $u$, $v$, $\sigma(v)$ and $\mu(v)$ have the same meaning as in Equation~\ref{spade}, $IN(\cdot)$ is the operation of instance normalization. $R\_SPADE$ is a generalization of $IN$. When the learned parameters $\sigma(v)$ and $\mu(v)$ converge to zero, $R\_SPADE$ can arrive at $IN$. Compared with Equation~\ref{spade}, $R\_SPADE$ has a residual structure that has been shown to facilitate the learning of network parameters~\cite{he2016deep}.
With the dynamic fine-tuning by $R\_SPADE$, the final depth features tend to retain the discrimination as much as possible while reducing the interference of their respective styles. Experiments in Section~\ref{experiment} demonstrate the effectiveness of MuSe-Net.

\subsection{Overview of MuSe-Net}\label{section:3}
The proposed Multiple-environment Self-adaptive Network (MuSe-Net) is illustrated in Figure~\ref{fig:2} (I). MuSe-Net consists of two branches: the multiple-environment style extraction branch and the self-adaptive feature extraction branch. These two branches have the same inputs. The multiple-environment style extraction branch is employed to extract features with different style information. Then these style features as the control data are fed into Residual SPADE to conduct a learnable transformation. 
Subsequently, the self-adaptive feature extraction branch employs the learned affine parameters of Residual SPADE to dynamically align the style information of inputs and pulls the content features with the same identity together. 

\textbf{The multiple-environment style extraction branch.} This branch has two components: a style encoder and a style classifier. The style encoder is intercepted from ResNet-50~\cite{he2016deep}. In particular, ResNet-50 contains four stages with repeated bottlenecks named stage1, stage2, stage3 and stage4. Following the analysis~\cite{pan2018two} that the style discrepancy is mostly preserved in shallow layers, we pick the first two stages of ResNet-50 as the style encoder to extract the style feature, as shown in Figure~\ref{fig:2} (II). The style classifier consists of a batch normalization layer (BN), a dropout layer (Dropout), and a fully-connected layer (FC). Given an aerial-view image $x_{j}^{i}$ (\ie{}, drone or satellite) of size $256 \times 256 \times 3$, we first utilize the style encoder to acquire the style feature $f_{j\_k}^{i}$ with the shape of $32 \times 32 \times 512$. Then we employ the average pooling layer to transform $f_{j\_k}^{i}$ into a 512-dim feature $g_{j\_k}^{i}$. In the end, we deploy the style classifier to predict the style of the input and utilize the cross-entropy loss to optimize this branch. The above process could be formulated as:
\begin{align}\label{extract1}
    f_{j\_k}^{i} &= E_{style}(x_{j}^{i}), 
\end{align}
\begin{align}\label{Avgpool}
    g_{j\_k}^{i} &= Avgpool(f_{j\_k}^{i}), 
\end{align}
\begin{align}\label{softmax1}
    p(y_{style}^{i}|x_{j}^{i}) &= softmax(\Phi_{s}(g_{j\_k}^{i})), 
    % \notag\\ &= \frac{exp(\Phi(g_{j\_k}^{i})[y_{style}^{i}])}{\sum_{k=0}^{K}exp(\Phi(g_{j\_k}^{i})[k]))},
\end{align}
\begin{align}\label{loss1}
    L_{style} &= \sum_{i,j}-log(p(y_{style}^{i}|x_{j}^{i})),
\end{align}
where $\Phi_{s}$ denotes the style classifier. $p(y_{style}^{i}|x_{j}^{i})$ is the predicted probability of $x_{j}^{i}$ belonging to the corresponding style label $y_{style}^{i}$. In Equation \ref{loss1}, we calculate the cross-entropy loss.
\par 
\textbf{The self-adaptive feature extraction branch.}
This branch consists of a content encoder with Residual SPADE embedded and an identity classifier. The content encoder is proposed relying on IBN-Net~\cite{pan2018two}. IBN-Net has the similar structure with ResNet-50. Specifically, the number of bottlenecks in stage1 of IBN-Net is 3. We embed Residual SPADE in the second and last bottlenecks of stage1 as the content encoder (see Figure~\ref{fig:2} (II)). Residual SPADE contains two convolutional layers, \ie, Conv\_w1 and Conv\_b1. One convolutional layer is employed to learn the scale, and another is for the bias (see Figure~\ref{fig:2} (III)). The identity classifier has the same component with the style classifier, \ie, a batch normalization layer (BN), a dropout layer (Dropout), and a fully-connected layer (FC). The content encoder accepts the same input $x_{j}^{i}$ with the style encoder, and is employed to extract the content feature $f_{j\_c}^{i}$ of size $16 \times 16 \times 2048$. Residual SPADE serves as an important role in the content encoder. In Residual SPADE, the style feature $f_{j\_k}^{i}$ is first interpolated to the same size as the activation of instance normalization (IN) in the content encoder. Then the interpolated feature is convolved to produce the scale and bias to modulate the activation of IN, and the modulation operation is shown in Figure~\ref{fig:2} (III-b). The extracted content feature is further transformed into a 2048-dim feature $g_{j\_c}^{i}$ by an average pooling layer. Finally, we harness the identity classification loss as the proxy target to force the content encoder to extract the discriminative feature, and this loss can be formulated as:
\begin{align}\label{softmax2}
    p(y_{ID}^{i}|x_{j}^{i}) &= softmax(\Phi_{ID}(g_{j\_c}^{i})), 
\end{align}
\begin{align}\label{loss2}
    L_{ID} &= \sum_{i,j}-log(p(y_{ID}^{i}|x_{j}^{i})),
\end{align}
where $\Phi_{ID}$ indicates the identity classifier. $p(y_{ID}^{i}|x_{j}^{i})$ is the predicted probability that $x_{j}^{i}$ belongs to the geo-tagged identity label $y_{ID}^{i}$.

\textbf{Optimization.} 
We train MuSe-Net by jointly employing the style loss $L_{style}$ and the identity loss $L_{ID}$:
\begin{align}\label{loss3}
    L_{total} &= L_{style} + L_{ID}.
\end{align}
The style loss forces features with different style information to stay apart, and the identity loss brings matching image pairs of the same geo-tag closer. Also, the identity loss serves the optimization of Residual SPADE.
\par
The pseudocode for MuSe-Net is shown as Algorithm~\ref{algorithm}.
% In the test phase, 
% When testing, we adapt the average pooling to process the content feature as the final image representation.
\begin{algorithm}[htp]
    \caption{PyTorch-style pseudocode for MuSe-Net.}
    \definecolor{codeblue}{rgb}{0.25,0.5,0.5}
    \lstset{
      basicstyle=\fontsize{7.2pt}{7.2pt}\ttfamily\bfseries,
      commentstyle=\fontsize{7.2pt}{7.2pt}\color{codeblue},
      keywordstyle=\fontsize{7.2pt}{7.2pt},
    }
\begin{lstlisting}[language=python, breaklines = true]
# model: MuSe_Net
# N: The batch size of one platform, \ie, drone or satellite.
# style_list: A list containing 10 transformations of the environmental style.
# x: The input image.
# X: One batch of input images.
# ys: The style label which belongs to [0, 10]. The style label of all satellite images is 0. The style labels of drone images are determined by the index of the style transformations in the style_list. For instance, the index of the rain style is 2, then the style label of a rain-style drone image is 3, \ie, index + 1.
# Ys: One batch of ys
# yi: The identity label of cross-platform geo-localization images.
# Yi: One batch of yi.
# CE: The cross entropy loss.
# _s: The satellite platform.
# _d: The drone platform.
# _sty: The images with environmental styles.
for data_s, data_d in loader_s, loader_d: # load a satellite batch and a drone batch with N samples, separately.
    X_s, Yi_s = data_s
    X_d, Yi_d = data_d
    X_d_sty = []
    Ys_d = []
    
    for x_d in X_d:
        generator, ys_d = random_select(style_list) # generator is a function of the style transformation, and ys_d is the corresponding style label.
        x_d_sty = generator(x_d)
        X_d_sty.append(x_d_sty)
        Ys_d.append(ys_d)
    X_s_sty = X_s
    Ys_s = [0] * N
    
    Yi_s',Yi_d',Ys_s',Ys_d' = model(X_s_sty, X_d_sty) # The outputs are the prediction vectors
    sty_loss = CE(Ys_s',Ys_s)+CE(Ys_d',Ys_d) # Equation (6)
    id_loss = CE(Yi_s',Yi_s)+CE(Yi_d',Yi_d) # Equation (8)
    loss = id_loss + sty_loss
    
    # optimization step
    loss.backward()
    optimizer.step()
        
\end{lstlisting}
\label{algorithm}
\end{algorithm}

\section{Experiments}\label{experiment}
We first introduce two datasets used for MuSe-Net and the evaluation protocol in Section~\ref{datasets}. Then we describe the implementation details in Section~\ref{implementation}. The comparison of results with different methods is provided in Section~\ref{results}, and the model analysis is followed in Section~\ref{ablation}.

\subsection{Datasets and Evaluation Protocol}\label{datasets}
We mainly train and evaluate the proposed method using the University-1652~\cite{zheng_university-1652_nodate} since it supports large-scale cross aerial-view images. We also verify the effectiveness of our method in SUES-200~\cite{zhu2023sues} and CVUSA~\cite{zhai_predicting_2017}.
% , which are  a street-to-satellite dataset.

\textbf{University-1652}~\cite{zheng_university-1652_nodate} is a newly-released dataset that focuses on the drone-based geo-localization. It consists of data from three different platforms, \ie{}, drones, satellites, and phone cameras.
% All of these data support two new tasks,
All of these data are collected from 1,652 buildings of 72 universities around the world. There are 54 drone-view images for one building in the dataset to guarantee that the drone-view data can cover rich information of the target, \eg{}, scale and viewpoint variants. With one satellite-view image for each building as a reference, the dataset also includes 5,580 street-view images. Due to the limited viewpoint of the phone camera, street-view images can not cover all facets of a target building. To make up this weakness as much as possible, 21,099 common-view images from Google Image are added to University-1652 as an extra training set. The training set includes 701 buildings of 33 universities, and another 951 buildings belonging to the rest 39 universities are contained in the test set. Universities in the training and test set are not overlapping. The dataset support two new tasks, 
\ie{}, drone-view target localization (Drone $\rightarrow$ Satellite) and drone navigation (Satellite $\rightarrow$ Drone). In the drone-view target localization task, the query set contains 37,855 drone-view images, and the gallery set includes 701 true-matched satellite-view images and 250 distractors. For the drone navigation task, with 701 satellite-view images as the query set, there are 37,855 true-matched drone-view images and 13,500 distractors in the gallery. Under this task, one query image has multiple correct matches in the gallery. It is clear that the drone-view target localization is a more challenging task than drone navigation since there is only one true-matched satellite-view image for a drone-view query. 

\textbf{SUES-200}~\cite{zhu2023sues} is a multi-height, multi-scene dataset which considers geo-localization between drone and satellite platforms. The dataset contains 200 scenes, and the drone-view images are recorded at four different heights: 150m, 200m, 250m, and 300m. At each height, one satellite-view scene corresponds to 50 drone-view images. It is worth noting that all drone-view images in SUES-200 are captured by a drone in real-world scenes.

\textbf{CVUSA}~\cite{zhai_predicting_2017} consists of image pairs from two viewpoints, \ie{}, the street view and the satellite view. Each viewpoint contains 35,532 images for training and 8,884 images for testing. It is worth noting that street-view images are panoramas, and all the street and satellite images are north aligned.

\textbf{Evaluation protocol.} The performance of our method is evaluated by the Recall@K (\textbf{R@K}) and the average precision (\textbf{AP}). \textbf{R@K} denotes the proportion of correctly localized images in the top-K list, and \textbf{R@1} is an important indicator. \textbf{AP} is equal to the area under the Precision-Recall curve.
Higher scores of \textbf{R@K} and \textbf{AP} indicate better performance of the network. 
\subsection{Implementation Details}~\label{implementation} 
The style encoder is intercepted from ResNet-50~\cite{he2016deep} and initialized using the pre-trained weights on ImageNet~\cite{li2009imagenet}. The kernel size of the convolutional layer in Residual SPADE is $3 \times 3$, and the kernel is initialized with \textit{normal initialization}. We employ weights of IBN-Net50-a~\cite{pan2018two} which is trained on ImageNet~\cite{li2009imagenet} to initialize the content encoder. Following~\cite{zheng_university-1652_nodate}, the stride of the second convolutional layer and the last down-sample layer of the first bottleneck in stage4 of the content encoder is modified from 2 to 1. We fix the size of input images to $256 \times 256$ pixels when training and inference. In training, we augment satellite-view images by employing random cropping and flipping. For drone-view images, we first apply the library of \textit{imgaug}~\cite{imgaug} to change the environmental style of images. For instance, we aim to generate an image in overexposure. We need to first convert a RGB image into the NumPy format using $img=np.array(img)$. Then we process the image in NumPy format using the following codes:
% \begin{equation}
% \begin{aligned}
%     aug &= iaa.MultiplyAndAddToBrightness(mul=1.6, add=(0, 30), seed=1992),
% \end{aligned}
% \end{equation}
% \begin{align}
%     img &= aug(image = img).
% \end{align} 
\definecolor{codeblue}{rgb}{0.25,0.5,0.5}
\lstset{
	basicstyle=\fontsize{8pt}{8pt}\ttfamily\bfseries,
	commentstyle=\fontsize{8pt}{8pt}\color{codeblue},
	keywordstyle=\fontsize{8pt}{8pt},
	xleftmargin=-5em,
	xrightmargin=4em
}
\begin{lstlisting}[language=python, breaklines = true]
	# img: The input image
	# img1: The input image in numpy format
	# img2: The output image with overexposure style
	# mul, add: Two parameters that control the brightness of the image
	import numpy as np
	import imgaug as iaa
	img1 = np.array(img)
	aug = iaa.MultiplyAndAddToBrightness(mul=1.6, add=(0,30), seed=1992)
	img2 = aug(image = img1).
\end{lstlisting}
The random cropping and flipping are subsequently performed to enhance these generated drone-view images. We adopt stochastic gradient descent (SGD) with
momentum 0.9 and weight decay 0.0005 to optimize our model. The mini-batch of training is set to 16 with 8 images for one platform. The initial learning rate is 0.005 for two classifiers and Residual SPADE, and 0.0005 for the rest layers. Our model is trained for 210 epochs, and the learning rate is decayed to its 0.1 and 0.01 at 120 and 180 epochs. During testing, the Euclidean distance is applied to measure the similarity between the query and candidate images in the gallery. We implement our code based on Pytorch~\cite{paszke2019pytorch}, and all experiments are conducted on one NVIDIA RTX 2080Ti GPU. Our model takes about 2 hours to train on University-1652. For one environmental condition, the testing time of Drone $\rightarrow$ Satellite and Satellite $\rightarrow$ Drone are 4 minutes, 52 seconds and 5 minutes, 44 seconds, respectively. 

\subsection{Comparisons with Competitive Methods}~\label{results}
% In experiments, we re-implement two methods as the competitive baselines using the same setting as our method. Both baselines include only the feature extraction branch containing a content encoder and an identity classifier. The backbone of the content encoder in baseline Zheng~\etal~\cite{zheng_university-1652_nodate} is ResNet-50, and in baseline IBN-Net~\cite{pan2018two} is IBN-Net50-a. 
\textbf{Results on University-1652.} University-1652~\cite{zheng_university-1652_nodate} supports two tasks: drone-view target localization (Drone $\rightarrow$ Satellite) and drone navigation (Satellite $\rightarrow$ Drone). We re-implement seven methods as competitive comparisons of our method on these two tasks. Seven comparison methods include only the feature extraction branch containing a content encoder and an identity classifier. Content encoders in seven comparison methods are VGG16~\cite{vgg16}, ResNet-50~\cite{he2016deep} (Zheng~\etal~\cite{zheng_university-1652_nodate}), ResNet-101~\cite{he2016deep}, DenseNet121~\cite{huang2017densely}, Swin-T~\cite{liu2021swin}, IBN-Net50-a (IBN-Net~\cite{pan2018two}), and LPN~\cite{wang2021each}. Both Swin-T~\cite{liu2021swin} and LPN~\cite{wang2021each} were published in 2021. Keeping the style of satellite-view images unchanged, the results of drone-view images in 10 different conditions are shown in Table~\ref{table:d2s}. In seven re-implemented methods of Drone $\rightarrow$ Satellite, LPN~\cite{wang2021each} explicitly considers the local information and obtains the best results. Excluding LPN, we observe that IBN-Net~\cite{pan2018two} has significantly improved geo-localization performance. Our method surpasses IBN-Net~\cite{pan2018two} in all environmental conditions. Specifically, when calculating the mean accuracy, our method improves the R@1 accuracy from $62.30\%$ to $65.15\%$ ($+2.85\%$) and the AP accuracy from $66.46\%$ to $69.16\%$ ($+2.70\%$). Meanwhile, our method also exceeds LPN~\cite{wang2021each}.
The Satellite $\rightarrow$ Drone is an easier task than Drone $\rightarrow$ Satellite. We first observe that even VGG16 can obtain higher performance of R@1 than the reported results of our method in Drone $\rightarrow$ Satellite. In Satellite $\rightarrow$ Drone, our method still keeps sufficient advantages over six comparison methods that do not utilize local information. In particular, our results in 10 different environmental conditions outperform all of the IBN-Net~\cite{pan2018two}, and the mean accuracy of R@1 increases from $82.27\%$ to $84.68\%$ ($+2.41\%$) and the mean value of AP raises from $63.36\%$ to $65.75\%$ ($+2.39\%$). Besides, our method is still competitive compared to LPN. The experimental results of two sub-tasks demonstrate two points. First, as the multiple-domain related method, IBN-Net~\cite{pan2018two} compared with other comparison methods with the same feature treatment can acquire a more robust representation containing less domain shift caused by different environmental styles. Second, our method based on IBN-Net learns the dynamic parameters to adaptively adjust the style information and can further improve the performance, as discussed in Section~\ref{section:2}. It is worth noting that our performance is still impacted by different environments, and some scenes have heavy invisibility, such as buildings in fog or dark. Our method minimizes the prediction discrepancy between the target weathers and normal cases. We intend to relieve the negative impact of the noisy environment to improve the robustness.

\textbf{Results on SUES-200.} 
The drone-view images on SUES-200 are divided into four groups according to the collected heights. We choose the hardest group, the drone-view images at 150m height, to conduct experiments. 
In experiments, the normal drone-view images are acquired in real environments, while other environmental images are synthesized on the basis of the normal situation. The results of our method are shown in Table~\ref{table:sues200}. 
We observe a similar phenomenon as for experiments on University-1652. We mainly compare our method with IBN-Net~\cite{pan2018two}. Our method improves the mean accuracy of R@1 from $39.58\%$ to $41.59\%$ ($+2.01\%$) and increases the mean value of AP from $46.23\%$ to $48.53\%$ ($+2.30\%$) in Drone $\rightarrow$ Satellite. In Satellite $\rightarrow$ Drone, our method raises the mean accuracy of R@1 from $52.25\%$ to $53.38\%$ ($+1.13\%$) and goes up the mean value of AP from $37.78\%$ to $39.20\%$ ($+1.42\%$). The effective performance improvement lays the foundation for geo-localization in realistic multiple environments.
\begin{table*}[tb]
\centering
%\small
\caption{
The R@1(\%) and AP(\%) accuracy for drone-view target localization task (Drone $\rightarrow$ Satellite) and drone navigation task (Satellite $\rightarrow$ Drone) on University-1652. In these two tasks, drone-view images hold 10 different environmental styles, and the style of satellite-view images is constant. The best results are in bold. 
}
\resizebox{\textwidth}{!}{
\begin{tabular}{l|cc|cc|cc|cc|cc|cc|cc|cc|cc|cc|cc}
\hline
\multicolumn{1}{c|}{\multirow{2}{*}{Method}} & \multicolumn{2}{c|}{Normal} & \multicolumn{2}{c|}{Fog} & \multicolumn{2}{c|}{Rain} & \multicolumn{2}{c|}{Snow} & \multicolumn{2}{c|}{\makecell[c]{Fog\\+Rain}} & \multicolumn{2}{c|}{\makecell[c]{Fog\\+Snow}} & \multicolumn{2}{c|}{\makecell[c]{Rain\\+Snow}} & \multicolumn{2}{c|}{Dark} & \multicolumn{2}{c|}{\makecell[c]{Over\\-exposure}} & \multicolumn{2}{c|}{Wind} & \multicolumn{2}{c}{\textbf{Mean~$\uparrow$}} \\ 
\cline{2-23}
                                        & R@1 & AP & R@1 & AP & R@1 & AP & R@1 & AP & R@1 & AP & R@1 & AP & R@1 & AP & R@1 & AP & R@1 & AP & R@1 & AP & R@1 & AP\\
\shline
\multicolumn{23}{c}{\textbf{Drone $\rightarrow$ Satellite}} \\
\shline
VGG16~\cite{vgg16} & 59.98 & 64.69 & 56.21 & 61.11 & 53.97 & 58.90 & 50.07 & 55.08 & 50.43 & 55.63 & 42.77 & 48.01 & 51.08 & 56.10 & 39.10 & 44.30 & 45.16 & 50.47 & 50.84 & 56.05 & 49.96 & 55.03  \\
Zheng~\etal~\cite{zheng_university-1652_nodate} & 67.83 & 71.74 & 60.97 & 65.23 & 60.29 & 64.61 & 55.58 & 60.09 & 54.75 & 59.40 & 44.85 & 49.78 & 57.61 & 62.03 & 39.70 & 44.65 & 51.85 & 56.75 & 58.28 & 62.83 & 55.17 & 59.71  \\
ResNet-101~\cite{he2016deep} & 70.07 & 73.04 & 63.87 & 68.22 & 63.34 & 67.59 & 59.75 & 64.15 & 57.45 & 62.12 & 48.31 & 53.28 & 60.25 & 64.68 & 46.12 & 51.02 & 56.34 & 61.23 & 62.13 & 66.63 & 58.76 & 63.29  \\
DenseNet121~\cite{huang2017densely} & 69.48 & 73.26 & 64.25 & 68.47 & 63.47 & 67.64 & 59.29 & 63.70 & 59.68 & 64.13 & 50.41 & 55.20 & 60.21 & 64.57 & 48.57 & 53.41 & 54.04 & 58.88 & 60.74 & 65.14 & 59.01 & 63.44  \\
Swin-T~\cite{liu2021swin} & 69.27 & 73.18 & 66.46 & 70.52 & 65.44 & 69.60 & 61.79 & 66.23 & 63.96 & 68.21 & \textbf{56.44} & \textbf{61.07} & 62.68 & 67.02 & 50.27 & 55.18 & 55.46 & 60.29 & 63.81 & 68.17 & 61.56 & 65.95  \\
IBN-Net~\cite{pan2018two} & 72.35 & 75.85 & 66.68 & 70.64 & 67.95 & 71.73 & 62.77 & 66.85 & 62.64 & 66.84 & 51.09 & 55.79 & 64.07 & 68.13 & 50.72 & 55.53 & 57.97 & 62.52 & 66.73 & 70.68 & 62.30 & 66.46  \\
LPN~\cite{wang2021each} & 74.33 & 77.60 & 69.31 & 72.95 & 67.96 & 71.72 & 64.90 & 68.85 & 64.51 & 68.52 & 54.16 & 58.73 & 65.38 & 69.29 & 53.68 & 58.10 & 60.90 & 65.27 & 66.46 & 70.35 & 64.16 & 68.14 \\
\hline
Ours & \textbf{74.48} & \textbf{77.83} & \textbf{69.47} & \textbf{73.24} & \textbf{70.55} & \textbf{74.14} & \textbf{65.72} & \textbf{69.70} & \textbf{65.59} & \textbf{69.64} & 54.69 & 59.24 & \textbf{66.64} & \textbf{70.55} & \textbf{53.85} & \textbf{58.49} & \textbf{61.05} & \textbf{65.51} & \textbf{69.45} & \textbf{73.22} & \textbf{65.15} & \textbf{69.16}  \\
\shline
\multicolumn{23}{c}{\textbf{Satellite $\rightarrow$ Drone}} \\
\shline
VGG16~\cite{vgg16} & 75.89 & 58.50 & 75.18 & 55.42 & 71.61 & 53.03 & 68.19 & 48.29 & 71.18 & 49.34 & 65.48 & 40.87 & 69.47 & 50.03 & 64.34 & 35.74 & 64.91 & 44.20 & 68.90 & 49.53 & 69.52 & 48.50  \\
Zheng~\etal~\cite{zheng_university-1652_nodate} & 83.45 & 67.94 & 79.60 & 61.12 & 77.60 & 59.73 & 73.18 & 55.07 & 75.89 & 54.45 & 70.76 & 43.26 & 74.75 & 56.44 & 69.47 & 39.25 & 72.18 & 51.91 & 76.46 & 57.59 & 75.33 & 54.68  \\
ResNet-101~\cite{he2016deep} & 85.73 & 71.79 & 82.45 & 66.46 & 81.46 & 65.68 & 79.74 & 61.72 & 79.74 & 60.59 & 74.75 & 50.31 & 80.17 & 62.61 & 75.32 & 45.37 & 79.60 & 58.21 & 82.31 & 64.67 & 80.13 & 60.74  \\
DenseNet121~\cite{huang2017densely} & 83.74 & 70.34 & 82.31 & 66.32 & 81.17 & 65.23 & 78.60 & 60.33 & 79.46 & 61.66 & 74.61 & 51.14 & 78.46 & 61.68 & 74.47 & 47.88 & 74.32 & 55.26 & 78.32 & 61.63 & 78.55 & 60.15  \\
Swin-T~\cite{liu2021swin} & 80.74 & 68.94 & 81.03 & 67.46 & 81.17 & 66.39 & 78.46 & 61.33 & 79.17 & 64.65 & 74.89 & 56.57 & 78.89 & 63.49 & 75.61 & 48.43 & 76.60 & 56.57 & 78.74 & 64.45 & 78.53 & 61.83  \\
IBN-Net~\cite{pan2018two} & 86.31 & 73.54 & 84.59 & 67.61 & 84.74 & 69.03 & 80.88 & 64.44 & 83.31 & 63.71 & 77.89 & 52.14 & 83.02 & 65.74 & 78.46 & 50.77 & 79.46 & 58.64 & 84.02 & 67.94 & 82.27 & 63.36 \\
LPN~\cite{wang2021each} & 87.02 & \textbf{75.19} & 86.16 & \textbf{71.34} & 83.88 & 69.49 & 82.88 & 65.39 & 84.59 & 66.28 & 79.60 & \textbf{55.19} & 84.17 & 66.26 & \textbf{82.88} & 52.05 & 81.03 & \textbf{62.24} & 84.14 & 67.35 & 83.64 & 65.08 \\
\hline
Ours & \textbf{88.02} & 75.10 & \textbf{87.87} & 69.85 & \textbf{87.73} & \textbf{71.12} & \textbf{83.74} & \textbf{66.52} & \textbf{85.02} & \textbf{67.78} & \textbf{80.88} & 54.26 & \textbf{84.88} & \textbf{67.75} & 80.74 & \textbf{53.01} & \textbf{81.60} & 62.09 & \textbf{86.31} & \textbf{70.03} & \textbf{84.68} & \textbf{65.75} \\
\hline
\end{tabular}}
% \vspace{-1cm}
\label{table:d2s}
\end{table*}
\par 

\begin{table*}[tb]
\centering
%\small
\caption{
The R@1(\%) and AP(\%) accuracy for drone-view target localization task (Drone $\rightarrow$ Satellite) and drone navigation task (Satellite $\rightarrow$ Drone) on SUES-200. In these two tasks, drone-view images from 150m height hold 10 different environmental styles, and the style of satellite-view images is constant. The best results are in bold. }
\resizebox{\textwidth}{!}{
\begin{tabular}{l|cc|cc|cc|cc|cc|cc|cc|cc|cc|cc|cc}
\hline
\multicolumn{1}{c|}{\multirow{2}{*}{Method}} & \multicolumn{2}{c|}{Normal} & \multicolumn{2}{c|}{Fog} & \multicolumn{2}{c|}{Rain} & \multicolumn{2}{c|}{Snow} & \multicolumn{2}{c|}{\makecell[c]{Fog\\+Rain}} & \multicolumn{2}{c|}{\makecell[c]{Fog\\+Snow}} & \multicolumn{2}{c|}{\makecell[c]{Rain\\+Snow}} & \multicolumn{2}{c|}{Dark} & \multicolumn{2}{c|}{\makecell[c]{Over\\-exposure}} & \multicolumn{2}{c|}{Wind} & \multicolumn{2}{c}{\textbf{Mean~$\uparrow$}} \\ 
\cline{2-23}
                                        & R@1 & AP & R@1 & AP & R@1 & AP & R@1 & AP & R@1 & AP & R@1 & AP & R@1 & AP & R@1 & AP & R@1 & AP & R@1 & AP & R@1 & AP\\
\shline
\multicolumn{23}{c}{\textbf{Drone $\rightarrow$ Satellite}} \\
\shline
Zheng~\etal~\cite{zheng_university-1652_nodate} & 41.28 & 48.01 & 36.92 & 43.17 & 39.38 & 45.41 & 31.22 & 38.05 & 32.58 & 38.55 & 20.55 & 25.84 & 34.08 & 40.61 & 35.15 & 41.43 & 27.67 & 33.55 & 35.17 & 41.67 & 33.40 & 39.63  \\
IBN-Net~\cite{pan2018two} & 48.55 & 54.91 & 43.83 & 50.21 & 42.88 & 49.54 & 37.78 & 44.78 & 39.50 & 45.77 & 24.82 & 31.47 & 39.62 & 46.55 & 39.20 & 46.41 & \textbf{34.83} & \textbf{41.39} & 44.78 & 51.23 & 39.58 & 46.23  \\
\hline
Ours & \textbf{50.05} & \textbf{57.26} & \textbf{46.90} & \textbf{53.69} & \textbf{45.12} & \textbf{51.99} & \textbf{43.97} & \textbf{50.92} & \textbf{40.42} & \textbf{47.17} & \textbf{25.40} & \textbf{31.74} & \textbf{43.45} & \textbf{50.52} & \textbf{39.72} & \textbf{47.01} & 32.75 & 39.81 & \textbf{48.15} & \textbf{55.14} & \textbf{41.59} & \textbf{48.53}  \\
\shline
\multicolumn{23}{c}{\textbf{Satellite $\rightarrow$ Drone}} \\
\shline
Zheng~\etal~\cite{zheng_university-1652_nodate} & 52.50 & 40.75 & 48.75 & 34.23 & 48.75 & 35.16 & 45.00 & 29.69 & 45.00 & 28.21 & \textbf{42.50} & 18.22 & 45.00 & 31.19 & 45.00 & 33.32 & 42.50 & 23.47 & 42.50 & 31.52 & 45.75 & 30.58  \\
IBN-Net~\cite{pan2018two} & 55.00 & 45.10 & 51.25 & 41.66 & \textbf{55.00} & 39.86 & 47.50 & 37.37 & \textbf{50.00} & 35.16 & 41.25 & 23.85 & 53.75 & 38.30 & \textbf{53.75} & \textbf{39.71} & 55.00 & \textbf{33.33} & 60.00 & 43.48 & 52.25 & 37.78 \\
\hline
Ours & \textbf{58.75} & \textbf{48.10} & \textbf{56.25} & \textbf{44.71} & 51.25 & \textbf{42.13} & \textbf{52.50} & \textbf{40.78} & 48.75 & \textbf{37.66} & 38.75 & \textbf{24.23} & \textbf{53.75} & \textbf{40.74} & 52.50 & 35.38 & \textbf{60.00} & 31.96 & \textbf{61.25} & \textbf{46.35} & \textbf{53.38} & \textbf{39.20} \\
\hline
\end{tabular}}
% \vspace{-1cm}
\label{table:sues200}
\end{table*}
\par 

\begin{figure}[htbp]
  \centering
  \includegraphics[width=0.9\linewidth]{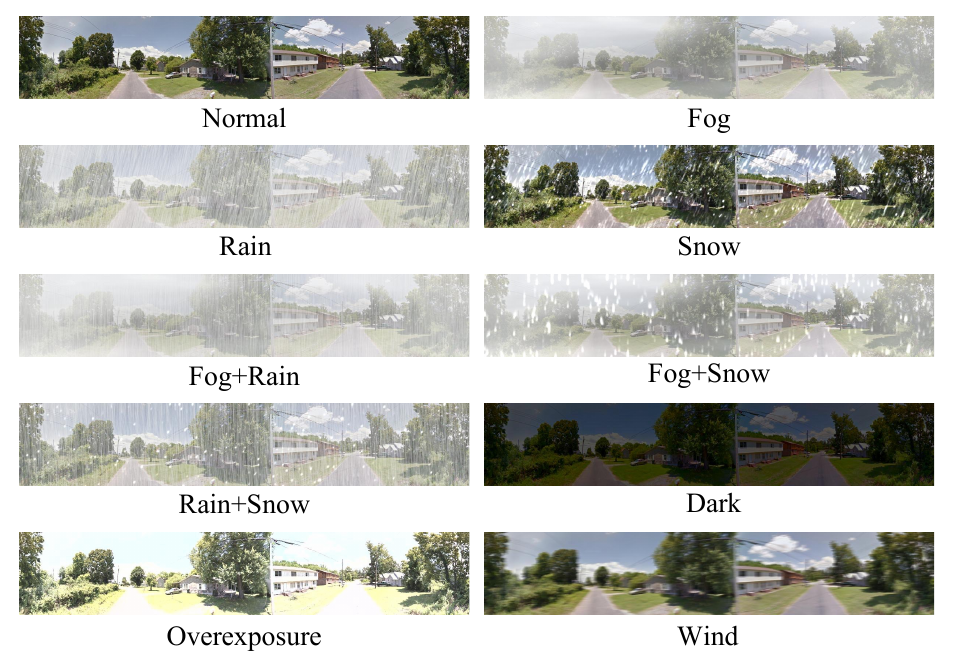}
%   \vspace{-.1in}
  \caption{Examples of a street-view panorama and its synthesized environments on CVUSA~\cite{zhai_predicting_2017}.}
  \label{fig:usa_envs}
\end{figure}

\textbf{Results on CVUSA.} 
Street-view and satellite-view images on CVUSA~\cite{zhai_predicting_2017} retain drastic appearance changes. In order to achieve cross-view images with a similar pattern, we follow~\cite{shi_spatial-aware_nodate,Shi_2020_CVPR} to pre-process satellite-view images before training and testing. Specially, we apply the polar transform to warp satellite images, which ensures that the appearance of satellite images is closer to street-view panoramas. In the multiple-environment setting, the satellite-view image is unchanged, and we only generate the multi-domain street-view images (see Figure~\ref{fig:usa_envs}). Results of our method compared with two competitive methods on CVUSA are detailed in Table~\ref{table:cvusa}. We could observe that our method obtains the increment in most environmental conditions than IBN-Net~\cite{pan2018two}. Meanwhile, the mean accuracy of R@1 goes up from $73.49\%$ to $75.00\%$ ($+1.51\%$), and the mean accuracy of AP boosts from $76.66\%$ to $78.04\%$ ($+1.38\%$).
\begin{table*}[tb]
\centering
%\small
\caption{
The R@1(\%) and AP(\%) performance of geo-localization on CVUSA. In this task, street-view images as queries have 10 different environmental style variations, while the style of satellite-view images in the gallery is constant. The best results are in bold. 
}
\resizebox{\textwidth}{!}{
\begin{tabular}{l|cc|cc|cc|cc|cc|cc|cc|cc|cc|cc|cc}
\hline
\multicolumn{1}{c|}{\multirow{2}{*}{Method}} & \multicolumn{2}{c|}{Normal} & \multicolumn{2}{c|}{Fog} & \multicolumn{2}{c|}{Rain} & \multicolumn{2}{c|}{Snow} & \multicolumn{2}{c|}{\makecell[c]{Fog\\+Rain}} & \multicolumn{2}{c|}{\makecell[c]{Fog\\+Snow}} & \multicolumn{2}{c|}{\makecell[c]{Rain\\+Snow}} & \multicolumn{2}{c|}{Dark} & \multicolumn{2}{c|}{\makecell[c]{Over\\-exposure}} & \multicolumn{2}{c|}{Wind} & \multicolumn{2}{c}{\textbf{Mean~$\uparrow$}} \\ 
\cline{2-23}
                                        & R@1 & AP & R@1 & AP & R@1 & AP & R@1 & AP & R@1 & AP & R@1 & AP & R@1 & AP & R@1 & AP & R@1 & AP & R@1 & AP & R@1 & AP\\
\shline
Zheng~\etal~\cite{zheng_university-1652_nodate} & 63.78 & 67.77 & 60.19 & 64.47 & 61.53 & 65.65 & 61.39 & 65.52 & 58.00 & 62.27 & 55.32 & 59.72 & 60.81 & 65.03 & 51.64 & 55.91 & 59.89 & 64.14 & 60.41 & 64.69 & 59.30 & 63.52 \\
IBN-Net~\cite{pan2018two} & 76.61 & 79.50 & 74.68 & 77.80 & 75.52 & 78.52 & 74.54 & 77.65 & 73.31 & 76.61 & 71.33 & 74.71 & 74.72 & 77.80 & 66.45 & 69.99 & 73.14 & 76.29 & 74.63 & 77.76 & 73.49 & 76.66 \\
\hline
Ours & \textbf{78.04} & \textbf{80.85} & \textbf{75.75} & \textbf{78.75} & \textbf{77.04} & \textbf{79.96} & \textbf{76.47} & \textbf{79.38} & \textbf{74.82} & \textbf{77.85} & \textbf{71.87} & \textbf{75.23} & \textbf{76.50} & \textbf{79.44} & \textbf{67.95} & \textbf{71.42} & \textbf{74.92} & \textbf{78.01} & \textbf{76.59} & \textbf{79.47} & \textbf{75.00} & \textbf{78.04} \\
\hline
\end{tabular}}
% \vspace{-1cm}
\label{table:cvusa}
\end{table*}
\par 
\textbf{Results on unseen weather.}
In the realistic scenario, the aerial-view geo-localization system can usually encounter the unseen weather. To explore whether MuSe-Net can cope with these weather conditions, especially the severe weather, we carry out experiments on mixing fog, rain and snow. Table~\ref{table:unseen} shows the experimental results. The proposed MuSe-Net on University-1652~\cite{zheng_university-1652_nodate} has achieved $44.10\%$ R@1 accuracy and $48.95\%$ AP for Drone $\rightarrow$ Satellite, and $75.32\%$ R@1 accuracy and $44.49\%$ AP for Satellite $\rightarrow$ Drone. The obtained R@1 accuracy and AP on CVUSA are $68.29\%$ and $71.81\%$, respectively. The superior performance compared with two competitive methods, \ie, Zheng~\etal{}~\cite{zheng_university-1652_nodate} and IBN-Net~\cite{pan2018two}, suggests that our method holds great potential to the unseen extreme weather. This character can also provide additional assurance of safe flight. 

\begin{table*}[tb]
\centering
\small
\caption{Results of retrieving in an unseen extreme weather, \ie, fog, rain and snow mixed.}
% \begin{center}
%\resizebox{\linewidth}{!}{
\begin{tabular}{l|cc|cc|cc}
\hline
\multicolumn{1}{c|}{\multirow{4}{*}{Method}} & \multicolumn{6}{c}{Fog + Rain + Snow} \\
\cline{2-7}
& \multicolumn{4}{c|}{University-1652} & \multicolumn{2}{c}{CVUSA} \\
\cline{2-7}
& \multicolumn{2}{c|}{Drone $\rightarrow$ Satellite} & \multicolumn{2}{c|}{Satellite $\rightarrow$ Drone} & \multicolumn{2}{c}{Street $\rightarrow$ Satellite}\\
& R@1 & AP & R@1 & AP & R@1 & AP\\
\shline
Zheng~\etal~\cite{zheng_university-1652_nodate} & 27.73 & 32.35 & 61.34 & 27.43 & 46.31 & 51.03 \\
IBN-Net~\cite{pan2018two} & 41.19 & 46.06 & 73.75 & 42.57 & 67.24 & 70.98\\
\hline
% Ours(SPADE) & 43.47 & 48.42 & 76.46 & 44.83 & - & \\
Ours & \textbf{44.10} & \textbf{48.95} & \textbf{75.32} & \textbf{44.49} & \textbf{68.29} & \textbf{71.81} \\
\hline
\end{tabular}
% \end{center}

\label{table:unseen}
\end{table*}

\begin{table*}[htp]
\centering
%\small
\caption{Ablation study of IBN-Net combined with SPADE and Residual SPADE, respectively. The best mean results are in bold.
}
\resizebox{\textwidth}{!}{
\begin{tabular}{l|cc|cc|cc|cc|cc|cc|cc|cc|cc|cc|cc}
\hline
\multicolumn{1}{c|}{\multirow{2}{*}{Method}} & \multicolumn{2}{c|}{Normal} & \multicolumn{2}{c|}{Fog} & \multicolumn{2}{c|}{Rain} & \multicolumn{2}{c|}{Snow} & \multicolumn{2}{c|}{\makecell[c]{Fog\\+Rain}} & \multicolumn{2}{c|}{\makecell[c]{Fog\\+Snow}} & \multicolumn{2}{c|}{\makecell[c]{Rain\\+Snow}} & \multicolumn{2}{c|}{Dark} & \multicolumn{2}{c|}{\makecell[c]{Over\\-exposure}} & \multicolumn{2}{c|}{Wind} & \multicolumn{2}{c}{\textbf{Mean}~$\uparrow$}\\ 
\cline{2-23}
                                        & R@1 & AP & R@1 & AP & R@1 & AP & R@1 & AP & R@1 & AP & R@1 & AP & R@1 & AP & R@1 & AP & R@1 & AP & R@1 & AP & R@1 & AP\\
\shline
IBN-Net~\cite{pan2018two} & 72.35 & 75.85 & 66.68 & 70.64 & 67.95 & 71.73 & 62.77 & 66.85 & 62.64 & 66.84 & 51.09 & 55.79 & 64.07 & 68.13 & 50.72 & 55.53 & 57.97 & 62.52 & 66.73 & 70.68 & 62.30 & 66.46  \\
+ SPADE~\cite{park2019semantic} & \textbf{74.51} & \textbf{77.88} & 67.70 & 71.65 & 68.60 & 72.55 & 64.61 & 68.76 & 63.93 & 68.20 & 53.80 & 58.51 & 65.76 & 69.86 & \textbf{53.91} & \textbf{58.62} & 58.94 & 63.55 & 68.41 & 72.36 & 64.02 & 68.19\\
\hline
+ Residual SPADE & 74.48 & 77.83 & \textbf{69.47} & \textbf{73.24} & \textbf{70.55} & \textbf{74.14} & \textbf{65.72} & \textbf{69.70} & \textbf{65.59} & \textbf{69.64} & \textbf{54.69} & \textbf{59.24} & \textbf{66.64} & \textbf{70.55} & 53.85 & 58.49 & \textbf{61.05} & \textbf{65.51} & \textbf{69.45} & \textbf{73.22} & \textbf{65.15} & \textbf{69.16} \\
\hline
\end{tabular}}
% \vspace{-1cm}
\label{table:sp vs re-sp}
\end{table*}

\begin{table*}[htp]
\centering
%\small
\caption{Results of embedding Residual SPADE in different IBN bottlenecks. \textbf{S} denotes the stage, and \textbf{B} stands for the bottleneck. For instance, \textbf{S1-B2,B3} means that the second and third IBN bottlenecks of stage1 are embedded with Residual SPADE. The best mean results are in bold.
}
\resizebox{\textwidth}{!}{
\begin{tabular}{l|cc|cc|cc|cc|cc|cc|cc|cc|cc|cc|cc}
\hline
\multicolumn{1}{c|}{\multirow{2}{*}{Method}} & \multicolumn{2}{c|}{Normal} & \multicolumn{2}{c|}{Fog} & \multicolumn{2}{c|}{Rain} & \multicolumn{2}{c|}{Snow} & \multicolumn{2}{c|}{\makecell[c]{Fog\\+Rain}} & \multicolumn{2}{c|}{\makecell[c]{Fog\\+Snow}} & \multicolumn{2}{c|}{\makecell[c]{Rain\\+Snow}} & \multicolumn{2}{c|}{Dark} & \multicolumn{2}{c|}{\makecell[c]{Over\\-exposure}} & \multicolumn{2}{c|}{Wind} & \multicolumn{2}{c}{\textbf{Mean}~$\uparrow$}\\ 
\cline{2-23}
                                        & R@1 & AP & R@1 & AP & R@1 & AP & R@1 & AP & R@1 & AP & R@1 & AP & R@1 & AP & R@1 & AP & R@1 & AP & R@1 & AP & R@1 & AP\\
\shline
S1-B1 & 73.76 & 77.13 & 69.02 & 72.79 & 69.26 & 72.98 & 64.44 & 68.48 & 64.57 & 68.67 & 54.12 & 58.17 & 66.27 & 70.18 & 52.71 & 57.34 & 59.93 & 64.42 & 67.33 & 71.21 & 64.14 & 68.14\\
S1-B2 & 73.74 & 77.20 & 68.69 & 72.56 & 69.17 & 72.93 & 65.51 & 69.54 & 64.53 & 68.71 & 54.63 & 59.26 & 66.61 & 70.52 & 53.96 & 58.57 & 59.53 & 64.08 & 68.45 & 72.26 & 64.48 & 68.56\\
S1-B3 & 74.28 & 77.69 & 68.91 & 72.79 & 68.74 & 72.61 & 64.09 & 68.29 & 64.58 & 68.77 & 54.05 & 58.78 & 66.00 & 70.01 & 52.79 & 57.56 & 59.33 & 63.95 & 68.05 & 72.05 & 64.08 & 68.25\\
S1-B1,B2 & 75.04 & 78.38 & 68.30 & 72.20 & 69.35 & 73.20 & 65.09 & 69.02 & 63.34 & 67.65 & 52.47 & 57.18 & 65.80 & 69.91 & 51.19 & 56.08 & 59.14 & 63.76 & 68.89 & 72.80 & 63.86 & 68.02\\
S1-B1,B3 & 73.99 & 77.46 & 68.28 & 72.30 & 69.33 & 73.17 & 64.63 & 68.92 & 64.05 & 68.39 & 52.95 & 57.83 & 65.77 & 69.94 & 51.40 & 56.33 & 58.49 & 63.22 & 67.88 & 71.95 & 63.68 & 67.95\\
S1-B2,B3 & 74.48 & 77.83 & 69.47 & 73.24 & 70.55 & 74.14 & 65.72 & 69.70 & 65.59 & 69.64 & 54.69 & 59.24 & 66.64 & 70.55 & 53.85 & 58.49 & 61.05 & 65.51 & 69.45 & 73.22 & \textbf{65.15} & \textbf{69.16} \\
S1-B1,B2,B3 & 74.85 & 78.15 & 68.12 & 72.02 & 68.94 & 72.77 & 65.91 & 69.95 & 62.81 & 67.06 & 53.57 & 58.35 & 66.46 & 70.49 & 52.58 & 57.41 & 59.27 & 63.87 & 68.57 & 72.45 & 64.11 & 68.25\\
\hline
\end{tabular}}
% \vspace{-1cm}
\label{table:btnk}
\end{table*}

\begin{table*}[htp]
\centering
%\small
\caption{Ablation study of two losses applying different weight ratios. The best mean results are in bold.
}
\resizebox{\textwidth}{!}{
\begin{tabular}{l|cc|cc|cc|cc|cc|cc|cc|cc|cc|cc|cc}
\hline
\multicolumn{1}{c|}{\multirow{2}{*}{$L_{ID}$ : $L_{style}$}} & \multicolumn{2}{c|}{Normal} & \multicolumn{2}{c|}{Fog} & \multicolumn{2}{c|}{Rain} & \multicolumn{2}{c|}{Snow} & \multicolumn{2}{c|}{\makecell[c]{Fog\\+Rain}} & \multicolumn{2}{c|}{\makecell[c]{Fog\\+Snow}} & \multicolumn{2}{c|}{\makecell[c]{Rain\\+Snow}} & \multicolumn{2}{c|}{Dark} & \multicolumn{2}{c|}{\makecell[c]{Over\\-exposure}} & \multicolumn{2}{c|}{Wind} & \multicolumn{2}{c}{\textbf{Mean}~$\uparrow$}\\ 
\cline{2-23}
                                        & R@1 & AP & R@1 & AP & R@1 & AP & R@1 & AP & R@1 & AP & R@1 & AP & R@1 & AP & R@1 & AP & R@1 & AP & R@1 & AP & R@1 & AP\\
\shline
1 : 0.5 & 73.67 & 77.10 & 67.47 & 71.42 & 68.75 & 72.59 & 65.11 & 69.15 & 63.54 & 67.74 & 54.47 & 59.22 & 65.93 & 69.95 & 52.32 & 56.98 & 58.45 & 63.02 & 67.97 & 71.85 & 63.77 & 67.90\\
1 : 1 & 74.48 & 77.83 & 69.47 & 73.24 & 70.55 & 74.14 & 65.72 & 69.70 & 65.59 & 69.64 & 54.69 & 59.24 & 66.64 & 70.55 & 53.85 & 58.49 & 61.05 & 65.51 & 69.45 & 73.22 & \textbf{65.15} & \textbf{69.16} \\
1 : 2 & 72.93 & 76.46 & 67.33 & 71.32 & 68.17 & 72.04 & 63.29 & 67.52 & 62.77 & 67.07 & 51.99 & 56.81 & 64.54 & 68.72 & 51.23 & 56.06 & 57.57 & 62.31 & 66.68 & 70.72 & 62.65 & 66.90\\
1 : 5 & 73.05 & 76.59 & 66.89 & 70.91 & 67.71 & 71.66 & 63.38 & 67.61 & 62.57 & 66.84 & 52.54 & 57.35 & 64.51 & 68.68 & 51.04 & 55.86 & 57.89 & 62.51 & 67.31 & 71.37 & 62.69 & 66.94\\
\hline
\end{tabular}}
% \vspace{-1cm}
\label{table:weights}
\end{table*}

\subsection{Model Analysis}\label{ablation}
We further analyze and discuss our model based on the multiple-environment drone-view target localization task (Drone $\rightarrow$ Satellite) in this section. 

\textbf{IBN-Net vs SPADE vs Residual SPADE.} 
As discussed in Section~\ref{section:2}, applying the adaptive adjustment strategy of style information (\ie, IBN-Net~\cite{pan2018two} + SPADE~\cite{park2019semantic}) can retain more discriminative features than IBN-Net when retrieval in different environments. Meanwhile, the proposed Residual SPADE with a residual structure is more effective than SPADE~\cite{park2019semantic}. The experiments are shown in Table~\ref{table:sp vs re-sp}. We observe first that IBN-Net combined with SPADE acquires a superior performance than only IBN-Net in all 10 different environments. Besides, Residual SPADE achieves higher results than SPADE in 8 environmental conditions. In the remaining environments (\ie, Normal and Dark), Residual SPADE also acquires similar results to SPADE. When considering the mean result, Residual SPADE goes up the R@1 accuracy from 64.02 to 65.15 ($+1.13\%$) and increases the AP value from 68.19 to 69.16 ($+0.97\%$). \textbf{The significant performance improvement further certifies the effectiveness of the adaptive adjustment strategy of style information and Residual SPADE.}

\textbf{Which bottleneck(s) embedding Residual SPADE is more effective?}
As mentioned in Section~\ref{section:3}, we embed Residual SPADE in the second and last bottlenecks of stage1. Other embedding options exist. Work~\cite{pan2018two} proves that the style difference mostly lies in the shallow layers. Following this finding, we select all three bottlenecks in stage1 for objects of our study. Specifically, we conduct experiments that embedding single, double or three Residual SPADE in these selected bottlenecks. Table~\ref{table:btnk} shows the details of experimental results. We observe first that deploying Residual SPADE in any bottlenecks yields higher mean results than IBN-Net~\cite{pan2018two} showed in Table~\ref{table:d2s}, which demonstrates the effectiveness of Residual SPADE. Then we compare results of deploying the single Residual SPADE in three different bottlenecks (\ie, the first three rows of Table~\ref{table:btnk}). Looking at mean results, embedding Residual SPADE in the second bottleneck ($S1-B2$) achieves the best performance. When employing two or three Residual SPADE in bottlenecks, we notice that combinations containing the first bottleneck, \ie, $S1-B1,B2$, $S1-B1,B3$ and $S1-B1,B2,B3$, achieve slightly lower mean results than $S1-B2,B3$. Finally, considering both the individual results in 10 conditions and the mean results, we choose $S1-B2,B3$ as the choice of our method.

\begin{figure}[htbp]
  \centering
  \includegraphics[width=0.9\linewidth]{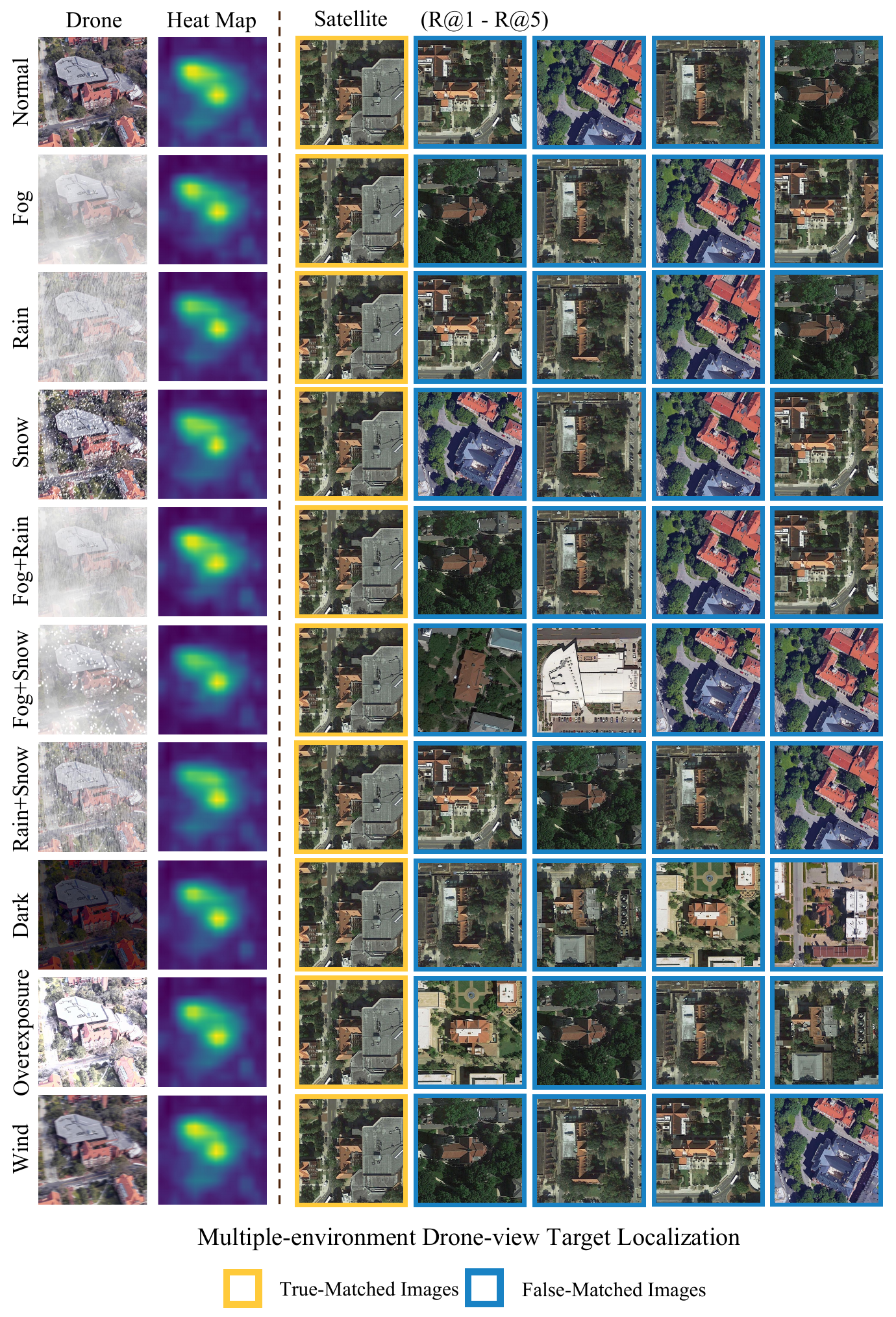}
%   \vspace{-.1in}
  \caption{Visualization of heatmaps generated by our method and Top-5 retrieval results for a drone-view image in different conditions. The true matches are in yellow boxes, and the false matches are displayed in blue boxes.}
  \label{heat&rank}
\end{figure}

\textbf{Effect of the loss weights.} MuSe-Net is a dual-branch network to disentangle the style and the localization information from drone-view images. The identity loss motivates the self-adaptive feature extraction branch to learn the \textbf{style-agnostic localization-aware} feature, while the style loss encourages the multiple-environment style extraction branch to learn the \textbf{style-aware localization-agnostic} feature. Since these two tasks are complementary and equally important, we set 1:1 as the default weight ratio of two losses. We further simply search other weight ratios of two losses, as shown in Table~\ref{table:weights}. 
The experimental results confirm that balanced identity/style losses give the best results.

\textbf{Qualitative result.}
As shown in Figure~\ref{heat&rank}, we visualise heatmaps and Top-5 retrieval results generated by our method in 10 different environmental conditions. Heatmaps show that our method can activate two regions of the geographic target. Another discovery is that there are subtle differences in the extent and brightness of the activated areas in 10 heatmaps. This phenomenon reflects from the side that results of geo-localization in 10 environmental conditions can exist difference. From the retrieval results shown, we observe that our model obtains the true match in the Top-1 yet the remaining retrieval results are inconsistent under 10 different conditions, which also indicates that the adjusted features still contain a few discrepancies.

\textbf{Model complexity.} 
We employ FLOPs and the parameter number to evaluate the model complexity of the proposed method and two baselines. FLOPs denotes the floating-point operations. The baseline methods of Zheng~\cite{zheng_university-1652_nodate} and IBN-Net~\cite{pan2018two} have similar complexity, \ie, $1.22 \times 10^{10}$ FLOPs and $48.43 million(M)$ parameter numbers. 
Our method inevitably yields a higher FLOPs ($1.70 \times 10^{10}$) and parameter number ($50.47 M$) since the designed multiple-environment style extraction branch. However, the growth rates of FLOPs and the parameter number between baselines and our method are $39.34\%$ and $4.21\%$, respectively. The lower costs of growth indicate that the complexity of our model is also acceptable. In addition, we also compare the parameter number of our method with ResNet-101~\cite{he2016deep} based method, Swin-T~\cite{liu2021swin} based method and LPN~\cite{wang2021each}. The parameter number of ResNet-101 based and Swin-T based method are $67.42 M$ and $53.14 M$, respectively. The parameter number of LPN is $52.66 M$. As shown in Table~\ref{table:d2s}, we notice that our performance is still competitive even compared to networks with a larger parameter number.

\section{Conclusion}
In this paper, we identify the challenge when employing aerial-view geo-localization in the real-world scenario where the weather and illumination changes. To reduce domain gaps of different environments, we propose an end-to-end learning network, called Multiple-environment Self-adaptive Network (MuSe-Net), to dynamically adjust the style difference for inputs with one geo-tag. MuSe-Net consists of two branches. One is a multiple-environment style extraction network for learning environment-related information. The other is a self-adaptive feature extraction network which integrates Residual SPADE into the content encoder to dynamically balance the environmental domain shift. To verify the effectiveness of MuSe-Net, we have evaluated the method in two drone-based geo-localization dataset (\ie, University-1652~\cite{zheng_university-1652_nodate} and SUES-200~\cite{zhu2023sues}) and achieved competitive performance. Besides, the proposed method also has acquired competitive results on one street-to-satellite dataset, \ie, CVUSA~\cite{zhai_predicting_2017}. In the future, we will continue to study the disentangled representation learning and further improve the performance of geo-localization in multiple environments.

\section*{Data Availability Statement}
Three datasets supporting the findings of this study are available with the permission of the dataset authors. The links to request these datasets are as follows. \\(1) University-1652 : \href{https://github.com/layumi/University1652-Baseline}{https://github.com/layumi/University1652-Baseline};\\ 
(2) SUES-200 : \href{https://github.com/Reza-Zhu/SUES-200-Benchmark}{https://github.com/Reza-Zhu/SUES-200-Benchmark}. \\
(3) CVUSA : \href{https://github.com/viibridges/crossnet}{https://github.com/viibridges/crossnet}.
\bibliography{mybibfile}

\end{document}